\documentclass[lettersize,journal]{IEEEtran}
\usepackage{amsmath,amsfonts}
\usepackage{algorithmic}
\usepackage{algorithm}
\usepackage{array}
\usepackage[caption=false,font=normalsize,labelfont=sf,textfont=sf]{subfig}
\usepackage{textcomp}
\usepackage{stfloats}
\usepackage{url}
\usepackage{verbatim}
\usepackage{graphicx}
\usepackage{cite}
\usepackage{threeparttable}
\usepackage{booktabs}
\usepackage{tabularx} % For adjustable-width columns
\usepackage{soul}

\usepackage{multirow}
\usepackage{makecell}

\usepackage{siunitx}

\usepackage{hyperref}

\usepackage[acronym]{glossaries}

\newacronym{ad}{AD}{Autonomous Driving}
\newacronym{il}{IL}{Imitation Learning}
\newacronym{rl}{RL}{Reinforcement Learning}
\newacronym{rlfd}{RLfD}{Reinforcement Learning from Demonstrations}
\newacronym{rlfold}{RLfOLD}{Reinforcement Learning from Online Demonstrations}
\newacronym{rlfp}{RLfP}{Reinforcement Learning from Pixels}
\newacronym{td}{TD}{Temporal Differences}
\newacronym{alix}{A-LIX}{Adaptive Local Signal Mixing}
\newacronym{ppo}{PPO}{Proximal Policy Optimization}
\newacronym{a3c}{A3C}{Asynchronous Advantage Actor Critic}
\newacronym{sac}{SAC}{Soft Actor-Critic}
\newacronym{ddpg}{DDPG}{Deep Deterministic Policy Gradient}
\newacronym{ema}{EMA}{Exponential Moving Average} 
\newacronym{bev}{BEV}{Bird’s Eye View}
\newacronym{irs}{$\textbf{IRS}$}{Infraction Rate Score}
\newacronym{ds}{$\textbf{DS}$}{Driving Score}
\newacronym{rc}{$\textbf{RC}$}{Route Completion}
\newacronym{ip}{$\textbf{IP}$}{Infraction Penalty}
\newacronym{ir}{$\textbf{IR}$}{Infraction Rate}
\newacronym{cnn}{CNN}{Convolution Neural Network}

\begin{document}

\title{PRIBOOT: A New Data-Driven Expert for Improved Driving Simulations}

\author{Daniel Coelho, Miguel Oliveira, Vítor Santos, and Antonio M. López
\thanks{Daniel Coelho is with the Department of Mechanical Engineering, University of Aveiro, 3810-193 Aveiro, Portugal, and with the Intelligent System Associate Laboratory (LASI), Institute of Electronics and Informatics Engineering of Aveiro (IEETA), University of Aveiro, 3810-193 Aveiro, Portugal (e-mail: danielsilveiracoelho@ua.pt).}
\thanks{Miguel Oliveira is with the Department of Mechanical Engineering, University of Aveiro, 3810-193 Aveiro, Portugal, and with the Intelligent System Associate Laboratory (LASI), Institute of Electronics and Informatics Engineering of Aveiro (IEETA), University of Aveiro, 3810-193 Aveiro, Portugal (e-mail: mriem@ua.pt)}
\thanks{Vítor Santos is with the Department of Mechanical Engineering, University of Aveiro, 3810-193 Aveiro, Portugal, and with the Intelligent System Associate Laboratory (LASI), Institute of Electronics and Informatics Engineering of Aveiro (IEETA), University of Aveiro, 3810-193 Aveiro, Portugal (e-mail: vitor@ua.pt)}
\thanks{Antonio M. López is with the Department of Computer Science, Computer Vision Center (CVC), Universitat Autonoma de Barcelona (UAB), Spain. (e-mail:antonio@cvc.uab.cat}
}

% \markboth{Journal of \LaTeX\ Class Files,~Vol.~14, No.~8, May~2023}%


\maketitle

\begin{abstract}

The development of \acrfull{ad} systems in simulated environments like CARLA is crucial for advancing real-world automotive technologies. To drive innovation, CARLA introduced Leaderboard 2.0, significantly more challenging than its predecessor. However, current \acrshort{ad} methods have struggled to achieve satisfactory outcomes due to a lack of sufficient ground truth data. Human driving logs provided by CARLA are insufficient, and previously successful expert agents like Autopilot and Roach, used for collecting datasets, have seen reduced effectiveness under these more demanding conditions. To overcome these data limitations, we introduce PRIBOOT, an expert agent that leverages limited human logs with privileged information. We have developed a novel \acrshort{bev} representation specifically tailored to meet the demands of this new benchmark and processed it as an RGB image to facilitate the application of transfer learning techniques, instead of using a set of masks. Additionally, we propose the \acrfull{irs}, a new evaluation metric designed to provide a more balanced assessment of driving performance over extended routes. PRIBOOT is the first model to achieve a \acrfull{rc} of 75\% in Leaderboard 2.0, along with a \acrfull{ds} and \acrshort{irs} of 20\% and 45\%, respectively. With PRIBOOT, researchers can now generate extensive datasets, potentially solving the data availability issues that have hindered progress in this benchmark.

\end{abstract}


\section{Introduction} \label{sec:introduction}

% Introduce Autonomous Driving: DONE!
% Problems of Real-World development: DONE!
% Importance of Simulators: DONE!
% Introduction of CARLA as the best open-source: DONE!
% Mention Leaderboard 1.0, and its high scores: DONE!
% Mention Leaderboard 2.0, and the low scores.  DONE!
% The potential reason for them. DONE!
% Mention our idea. Bootstrapping with privileged information. Decreasing the complexity of the problem.
% List the 3 contributions.

\acrfull{ad} is a key technological advancement with the potential to transform transportation, improve road safety, and redefine urban environments \cite{andersson2019benefits, 9832636}. Despite its potential, developing fully autonomous vehicles involves significant challenges. These include integrating diverse sensors, processing complex data, making real-time decisions, and addressing ethical issues. Such vehicles must operate reliably in unpredictable conditions, requiring advanced systems capable of handling a wide range of scenarios \cite{dosovitskiy2017carla}. Real-world testing of autonomous vehicles, while necessary, is often expensive, risky, and encumbered by ethical dilemmas.

Simulations serve as a critical complement to real-world testing, providing a safe and controlled environment that replicates complex real-world scenarios without the associated costs and risks \cite{li2024choose, dosovitskiy2017carla}. This enhances the development of autonomous driving technologies by allowing preliminary testing and refinement in simulations, reserving real-world trials for the final stages of development. Moreover, in these simulated environments, it is possible to leverage privileged information, otherwise not available in the real-world, to create expert systems that can provide demonstrations, further enriching the development process.

Among various open-source \acrshort{ad} simulators, CARLA \cite{dosovitskiy2017carla} is often listed as the premier choice \cite{kaur2021survey, li2024choose}. CARLA offers a suite of essential features for realistic and effective simulation of driving scenarios. These include comprehensive environmental conditions, detailed vehicle models, and a wide array of sensors, making it an ideal platform for advanced \acrshort{ad} research and development. 

\IEEEpubidadjcol

To accelerate innovation, CARLA introduced the CARLA Leaderboard 1.0\footnote{https://leaderboard.carla.org/\#leaderboard-10} benchmark, designed to assess the driving proficiency of autonomous agents within realistic traffic scenarios.
Despite the complex scenarios presented in the benchmark, various methods such as ReasonNet \cite{shao2023reasonnet}, InterFuser \cite{shao2023safety}, and TCP \cite{wu2022trajectory} have consistently reported high performance over the years. Notably, the CARLA Autopilot, a rule-based agent, achieved near-perfect performance. This underscores the benchmark's capacity to be effectively mastered using current technologies.
Building on this foundation, CARLA Leaderboard 2.0\footnote{https://leaderboard.carla.org/} introduces even more complex and challenging scenarios, such as obstacles in the lane and parking exits. These novel scenarios significantly increase the difficulty level, challenging the limits of existing autonomous driving systems. To this date, all approaches tested on the Leaderboard 2.0 benchmark have shown very poor performance, with the highest \acrfull{rc} reaching 15\%, and the highest \acrshort{ds} reaching 1\% \cite{zhang2024analysis}. We believe that the primary reason for this notable decline in performance can be attributed to the insufficiency of available training data. CARLA provides a set of human driving logs from a few route scenarios, but these are insufficient for training models that rely on sensor information. In Leaderboard 1.0, researchers could leverage online experts like CARLA Autopilot or Roach \cite{zhang2021end} to generate demonstrations. 
However, in Leaderboard 2.0, these experts are either markedly less effective or completely ineffective, as we will show in Section \ref{sec:related_work}.

This paper presents a method to address the challenges posed by the limited training data availability. The driving logs from CARLA, while insufficient alone for models requiring sensor inputs, become significantly more useful when combined with privileged information from the simulator, specifically \acrfull{bev}. This integration effectively simplifies the complexity of the benchmark. Employing this strategy, we apply \acrlong{il} techniques to develop PRIBOOT (\textbf{Pr}ivileged \textbf{I}nformation \textbf{Boot}strapping), an expert agent capable of navigating the demanding scenarios presented in Leaderboard 2.0. PRIBOOT utilizes privileged information to master these scenarios, subsequently enabling the generation of extensive datasets or providing online demonstrations. Although PRIBOOT was designed to address the challenges of Leaderboard 2.0, it is also applicable to any other CARLA benchmark.

Overall, we summarize our main contributions as follows:

\begin{itemize}
    \item Introduce PRIBOOT, an expert agent that effectively leverages privileged information and limited data for model training, marking the first instance of achieving significant performance milestones on the CARLA Leaderboard 2.0;
    \item Develop a tailored \acrfull{bev} representation to effectively address the complex driving scenarios encountered in CARLA Leaderboard 2.0.
    \item Process the \acrshort{bev} as an RGB image rather than a set of masks. This facilitates the application of transfer learning techniques, which significantly enhance model performance and efficiency, particularly in the context of limited data availability;
    \item Introduce \acrfull{irs}, a novel evaluation metric that considers infractions per kilometer rather than the total number of infractions. This metric is designed to complement the \acrfull{ds} by providing a more detailed assessment of driving behavior over long routes.
\end{itemize}

The source code of PRIBOOT is available at \href{https://github.com/DanielCoelho112/priboot}{https://github.com/DanielCoelho112/priboot}.
%--------------------------------------------------------------------------------------
%--------------------------------------------------------------------------------------
\section{Related Work} \label{sec:related_work}
%--------------------------------------------------------------------------------------
%--------------------------------------------------------------------------------------

This section is divided into two topics: the application of expert agents in \acrshort{ad}, and an overview of all expert agents utilized in CARLA.

\subsection{Application of Experts in \acrlong{ad}}

In recent years, the field of \acrshort{ad} has seen significant advancements through the application of online experts \cite{pan2017agile,zhang2021end,jia2023driveadapter,wu2022trajectory}. 
A notable example of this is showcased in \cite{pan2017agile}, where the utility of online experts is demonstrated in real-world, high-speed off-road driving scenarios. In their approach, an initial expert system equipped with expensive sensors is developed using a combination of hand-engineered components. This expert system then serves as a reference model, providing high-quality driving demonstrations to train a student model, which operates using more affordable sensors. 

Building on the foundational use of online experts, the transfer of knowledge from the expert to the student model can be accomplished through various methodologies. One prevalent method involves the creation of offline datasets, which are subsequently employed for offline \acrfull{il} \cite{codevilla2019exploring,shao2023reasonnet}  or \acrfull{rlfd} \cite{chekroun2023gri}. These approaches are particularly valuable in scenarios where direct interaction with the environment is either too costly or filled with risks. However, a significant challenge with using offline datasets is the potential for a distribution shift \cite{ross2011reduction}. To address the issue of distribution shift, an alternative strategy is online \acrshort{il} \cite{chen2020learning}, where the student actively explores the environment while the teacher provides on-demand supervision. This method helps to align the student's learning experience more closely with the actual operational environment, thereby reducing the impact of distribution shift. Nevertheless, this approach still relies heavily on the quality of the data provided by the expert. If the expert's behavior is not optimal, the student is likely to inherit these imperfections \cite{zhang2021end}.
To further refine this process and overcome the limitations of potentially suboptimal expert data, another innovative approach is \acrfull{rlfold} \cite{coelho2024rlfold}. This technique merges the benefits of Online \acrshort{il} with the principles of \acrshort{rl}, tackling both the issue of distribution shift and the challenge of learning from a suboptimal expert.

\subsection{Experts in CARLA}

CARLA incorporates a built-in expert system known as Autopilot, which relies on a series of handcrafted rules that utilize the internal state of the simulator for navigation \cite{dosovitskiy2017carla}. In Leaderboard 1.0, Autopilot demonstrated commendable performance, contributing significantly to data collection for top-ranked methods such as ReasonNet \cite{shao2023reasonnet}, which relies on datasets generated by this expert. However, the transition to Leaderboard 2.0 reveals a stark contrast in the efficacy of the Autopilot system. As detailed in Section \ref{sec:experiments}, the performance of Autopilot is markedly diminished in the more demanding scenarios of this updated benchmark. The primary challenge lies in the inherent limitations of a rule-based navigation framework, which struggles to adapt to the complex and dynamic driving conditions presented in Leaderboard 2.0, such as yielding to emergency vehicles or overtaking obstacles in the lane.

While Autopilot has shown competent performance in earlier benchmarks, the adoption of learning-based experts presents distinct advantages \cite{chen2020learning,zhao2021sam,zhang2021end}. These methods usually decouple the perception from planning, which simplifies the training process. Typically, such methods leverage privileged information from the simulator to bypass the need for complex perception systems, focusing instead on training the planning module directly. For example, LBC \cite{chen2020learning} and SAM \cite{zhao2021sam} replace the perception module with simulator-derived privileged information, and then train the planning component using \acrshort{il} based on demonstrations provided by the Autopilot. To ensure effective knowledge transfer from the expert to the student, LBC aims to minimize the output differences between them, whereas SAM focuses on aligning the latent representations of both models. These learning-based experts have been assessed in straightforward benchmarks, such as the NoCrash benchmark \cite{codevilla2019exploring}. This benchmark, with its relatively simple navigation and collision avoidance tasks, represents a significantly lesser challenge than even the earlier Leaderboard 1.0, and far less demanding than the complexities encountered in Leaderboard 2.0.

\begin{figure*}[t]
    \centering
    \subfloat[]{%
      \includegraphics[width=0.49\textwidth]{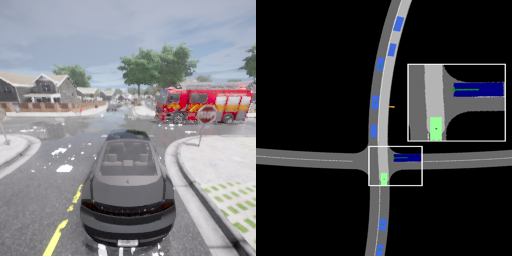}
      \label{fig:sub1}
    }
    \hfill
    \subfloat[]{%
      \includegraphics[width=0.49\textwidth]{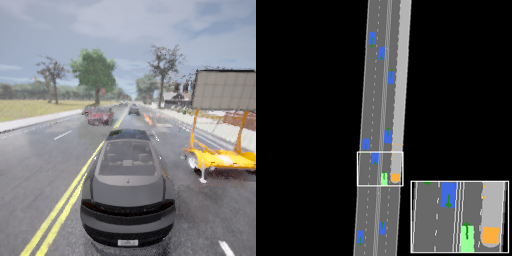}
      \label{fig:sub2}
    }
    \caption{\acrshort{bev} used in PRIBOOT. This representation was built upon Roach and LBC, with critical adaptations to tailor it for the complexities of Leaderboard 2.0. (a) Differentiates emergency vehicles (dark blue) from regular vehicles (blue). (b) Introduces an additional class for construction objects, illustrated in orange. 
    Additionally, both images depict a simple method to represent motion with directional arrows, illustrated in green. The images with a white frame provide a zoomed-in view to highlight specific details.}
    \label{fig:bev-adaptations}
\end{figure*}

More recently, the Roach expert \cite{zhang2021end} was introduced and has since become the most utilized expert in Leaderboard 1.0 \cite{wu2022trajectory,xiao2023scaling}. Roach processes inputs using a \acrfull{bev} image that encapsulates roads, lanes, routes, vehicles, pedestrians, traffic lights, and stop signs. This information is then processed using a model-free \acrfull{rl} algorithm to generate vehicle control commands. While Roach has demonstrated impressive results in Leaderboard 1.0, its applicability to Leaderboard 2.0 is questionable without significant modifications.
Several challenges hinder the transition of Roach to the more demanding Leaderboard 2.0. Firstly, their \acrshort{bev} implementation struggles with scalability issues in the larger towns of Leaderboard 2.0, primarily due to memory constraints when computing the cache for the roads and lanes. Secondly, the existing classes in the \acrshort{bev} representation fall short in capturing complex new scenarios introduced in the updated leaderboard, such as construction zones or the presence of emergency vehicles. 
Lastly, there is uncertainty regarding the effectiveness of Roach’s model-free \acrshort{rl} approach when faced with the heightened complexity and dynamic requirements of Leaderboard 2.0. It is important to note that Roach was trained for about one week on an Nvidia RTX 2080 Ti to achieve its results on Leaderboard 1.0. Considering the increased difficulty and complexity of scenarios in Leaderboard 2.0, adapting and retraining Roach could potentially require significantly more time, further complicating its deployment in this new benchmark.

Recognizing the limitations of existing expert agents for Leaderboard 2.0, CARLA has made available a set of driving logs that showcase human-driven routes in various scenarios. However, these logs alone do not suffice to train a system capable of processing sensor inputs and generating control commands. In response, and inspired by the approaches of LBC and Roach, we propose PRIBOOT, a method that simplifies the perception component by employing a \acrfull{bev} as the primary input. 
However, instead of using \acrshort{bev} as independent mask channels for training a \acrshort{cnn} from scratch, PRIBOOT converts the mask into an RGB image and leverages transfer learning techniques using pre-trained networks from the ImageNet dataset \cite{5206848}. This adaptation is crucial, particularly given the limited data available.

%--------------------------------------------------------------------------------------
%--------------------------------------------------------------------------------------
\section{Method}
%--------------------------------------------------------------------------------------
%--------------------------------------------------------------------------------------

PRIBOOT (\textbf{Pr}ivileged \textbf{I}nformation \textbf{Boot}strapping) leverages the limited logs available in Leaderboard 2.0 to establish the first expert agent capable of achieving satisfactory results within this demanding benchmark, as we show in Section \ref{sec:experiments}. The development of PRIBOOT was structured in two phases: First, we focused on generating the most effective input representation tailored to the unique challenges of Leaderboard 2.0, as detailed in Section \ref{sec:bev}. Following this, we designed and implemented a neural network architecture that is specifically optimized for handling the constraints of limited data, described in Section \ref{sec:arch}.

\begin{figure*}[t]
\centering
% \includesvg[width=2\columnwidth]{imgs/iccv_2023_network_v3}
\includegraphics[width=1.5\columnwidth]{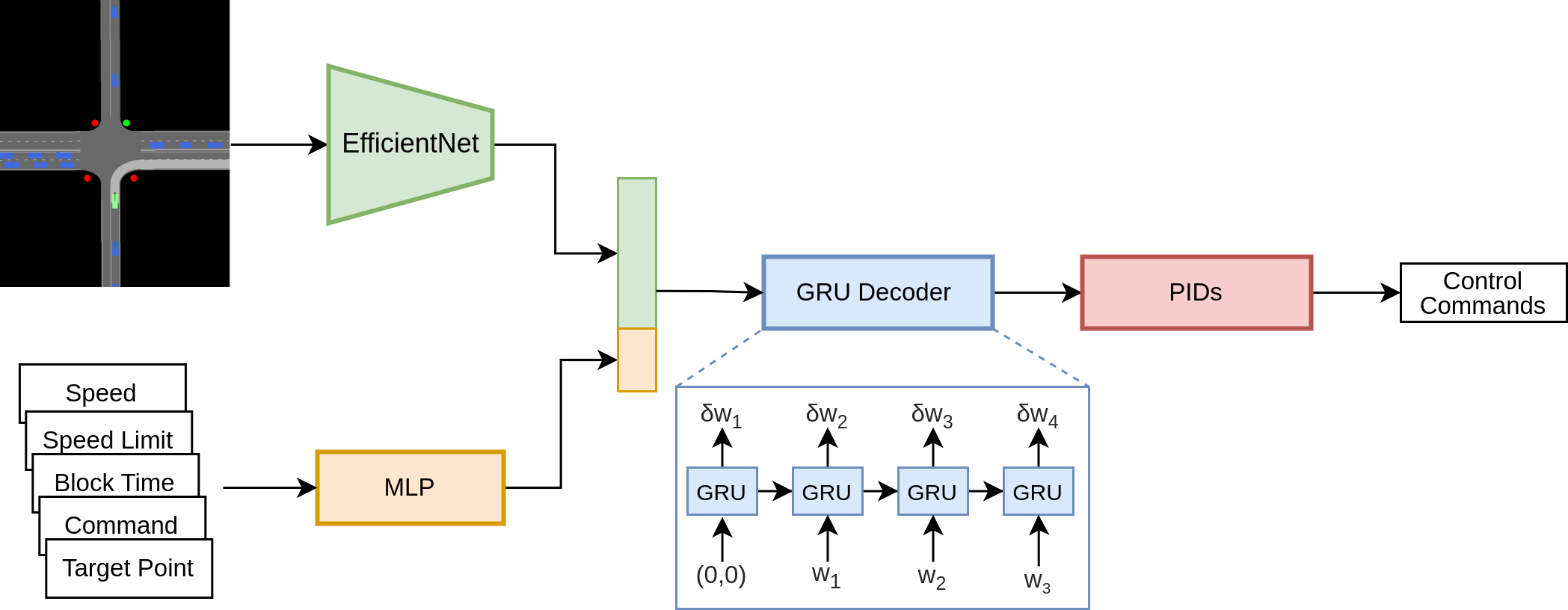}
\caption[Architecture of PRIBOOT]{Architecture of PRIBOOT. The system receives two types of inputs: a \acrshort{bev} image and a vector of vehicle measurements. These inputs are processed independently— the BEV through a pretrained EfficientNet model and the vehicle measurements via a MLP. The resultant feature vectors from both models are concatenated to form a comprehensive feature vector, which is then fed into a GRU-based waypoint decoder, similar to the approach used by Transfuser \cite{chitta2022transfuser}. The final stage involves processing the waypoints through both longitudinal and lateral PID controllers to generate the vehicle control commands.}
\label{fig:architecture}
\end{figure*}

\subsection{Generation of \acrlong{bev}} \label{sec:bev}

Building on the approach used by Roach \cite{zhang2021end} and LBC \cite{chen2020learning}, we employ a \acrshort{bev} to model the environment. However, adaptations were necessary to tailor it for the complexities of Leaderboard 2.0. Roach's and LBC's \acrshort{bev} include various classes such as roads, desired routes, lane boundaries, vehicles, pedestrians, traffic lights, and stop signs. While these classes were adequate for Leaderboard 1.0, they proved insufficient for the expanded scope of Leaderboard 2.0. Our enhancements to the \acrshort{bev} are outlined below:

\paragraph{Scalable Cache}
Current \acrshort{bev} approaches utilize a caching mechanism to store the roads and lanes for the entire town, a process completed once per town to facilitate real-time \acrshort{bev} generation. However, applying this method to the larger towns in Leaderboard 2.0 caused memory overflows due to the use of Pygame. We addressed this by adopting a more efficient caching technique inspired by deepsense.ai\footnote{https://github.com/deepsense-ai/carla-birdeye-view}, implementing the cache with NumPy for enhanced performance.

\paragraph{Decomposition of the Vehicles Class}
In current \acrshort{bev} representations, all vehicle types are aggregated under a single class. We refined this by segmenting the Vehicles class into three distinct categories: Bikes, Emergency Vehicles, and Regular Vehicles. This differentiation is crucial as the driving behavior varies significantly based on the type of nearby vehicle, especially in emergency situations (see Figure \ref{fig:sub1}).

\paragraph{Simplified Motion Representation}
Roach's \acrshort{bev} uses multiple temporal masks to capture movement, which increases significantly the computational load. Instead, we introduced a single additional mask featuring an arrow for each actor, as illustrated with green arrows in Figure \ref{fig:bev-adaptations}. This arrow indicates both the direction (orientation) and the speed (length) of the actor, simplifying the representation while reducing memory and computational demands.

\paragraph{Incorporation of a New Class}
Leaderboard 2.0 introduces scenarios requiring interaction with new environmental elements not covered by existing classes. For instance, construction zones that necessitate slight route deviations were not previously accounted for. To accommodate this, we added a new class named Construction, which encompasses all pertinent elements like traffic cones and street barriers, represented in orange in \ref{fig:sub2}.

\paragraph{RGB Format Instead of Masks}
Roach and LBC process the \acrshort{bev} using independent mask channels, requiring the training of a \acrshort{cnn} from scratch. Given the limited data in Leaderboard 2.0, we found that converting these masks into an RGB format and utilizing pre-trained visual encoders not only saves training time but also enhances the model's performance and efficiency.

\subsection{Architecture} \label{sec:arch}

The architecture of PRIBOOT is depicted in Figure 2. Our system takes as input a \acrshort{bev} image and a vector of vehicle measurements. The vehicle measurement vector encompasses several key parameters: current speed and the road speed limit, block time, a target point, and a navigation command. The "block time" parameter denotes the duration during which the vehicle has been stationary, aiding the system in determining whether to overtake or maintain its position due to typical traffic conditions. The "target point" is a waypoint located 30 meters ahead on the desired trajectory provided by a global planner, and the "navigation command" provides high-level directional indication from the global planner, encoded as a one-hot vector.

We utilize an EfficientNet \cite{tan2019efficientnet} for processing the BEV image and an MLP for handling vehicle measurements. Given the constraint of limited available data, we adopt transfer learning by employing the pretraining of EfficientNet with the ImageNet dataset. Subsequently, the feature vectors extracted from both the EfficientNet and the MLP are merged and inputted into an autoregressive GRU decoder. This decoder is tasked with predicting the subsequent T=4 waypoints $\{w_t\}^{T}_{t=1}$ within the ego-vehicle coordinate framework, drawing inspiration from the methodology applied in Transfuser \cite{chitta2022transfuser}.

To convert the predicted waypoints into control commands, we employ two PID controllers—one for lateral control and another for longitudinal control—following methodologies from \cite{chen2020learning, chitta2022transfuser}. The longitudinal controller uses the magnitude of the average vector between consecutive waypoints, while the lateral controller relies on their orientation. Additionally, similar to Transfuser, we integrate a creeping behavior and a safety heuristic mechanism utilizing information from the simulator.

This system is trained end-to-end using an $L_1$ loss between the predicted waypoints and the ground truth waypoints from the human logs. Let $w^{*}_{t}$ represent the ground truth waypoint at timestep $t$, the the loss function is defined as:

\begin{equation}
    \mathcal{L} = \sum_{t=1}^{T} \|w_t - w_{t}^{*}\|_1.
\end{equation}

\begin{figure}[t]
    \centering
    \subfloat[]{%
      \includegraphics[width=0.24\textwidth]{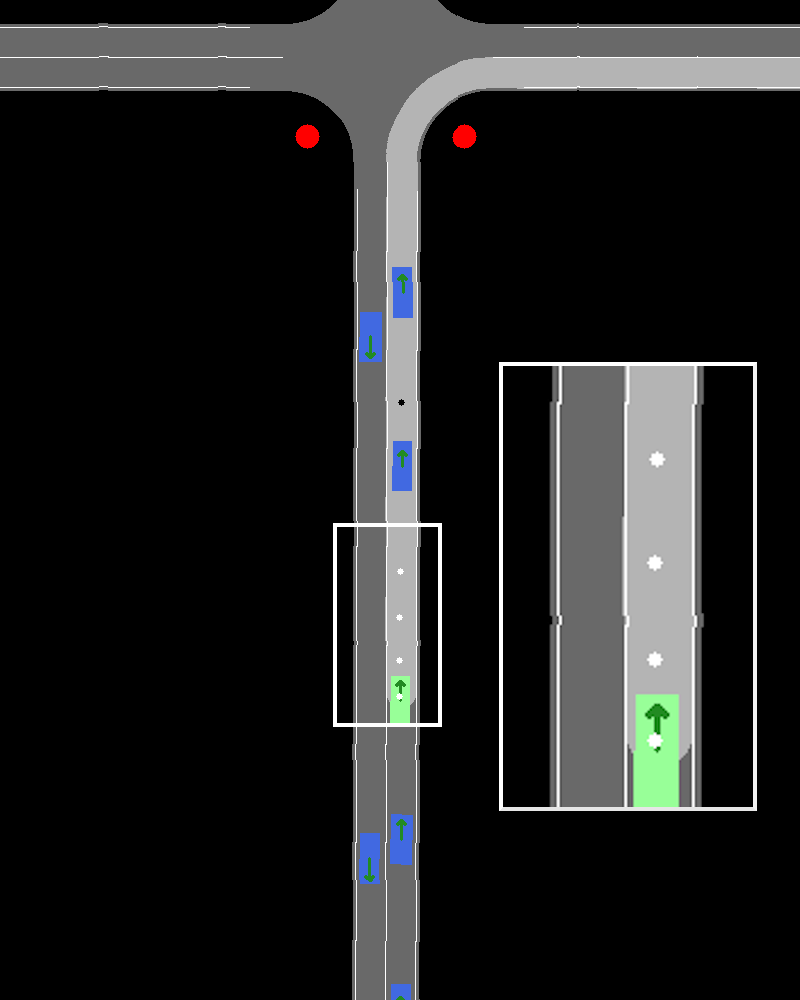}
      \label{fig:no_aug}
    }
    % \hfill
    \subfloat[]{%
      \includegraphics[width=0.24\textwidth]{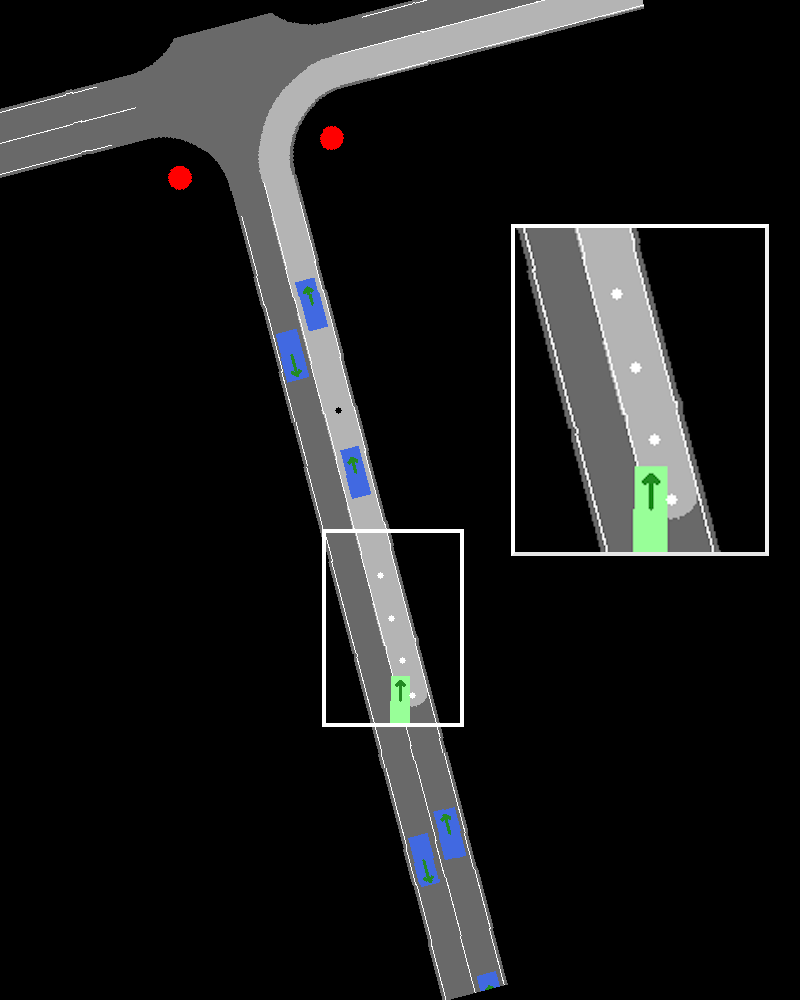}
      \label{fig:aug}
    }
    \caption{Data augmentation techniques used to expose the agent to a broader range of driving scenarios. In these illustrations, white dots indicate the future waypoints that were followed by the human driver, and the black dot represents the target point. (a) Displays a sample with no augmentation, showing the standard scenario. (b) Shows a sample where both translation and rotation augmentations have been applied to the ego vehicle, illustrating a situation where the agent needs to recover to the center of the lane.}

    \label{fig:bev-aug}
\end{figure}

The human demonstrations typically exhibit minimal deviation from the center of the lane, resulting in noise-free data. However, this adherence to the centerline causes a distribution shift between the training distribution and inference distribution. During inference, due to planning or controller inaccuracies, the agent may find itself in scenarios that deviate from the center of the lane. These instances are encountered as out-of-distribution events, presenting difficulties for the agent to navigate. To address this issue, inspired by the LBC approach, we use data augmentation techniques regarding the position and orientation of the vehicle. By varying the position and orientation of the vehicle, we expose the agent to diverse configurations, enabling it to learn effective recovery strategies. Figure \ref{fig:bev-aug} provides a visual representation of this augmentation process.

%--------------------------------------------------------------------------------------
%--------------------------------------------------------------------------------------
\section{Experiments} \label{sec:experiments}
%--------------------------------------------------------------------------------------
%--------------------------------------------------------------------------------------

This section starts with an overview of the setup used for collecting the experiments, followed by a comparative analysis of expert agents. It concludes with a presentation of an ablation study.

\subsection{Setup}

\subsubsection{Benchmark}

CARLA Leaderboard 2.0 builds upon CARLA Leaderboard 1.0, increasing the complexity of the benchmark in three distinct ways: a) by extending the route lengths approximately tenfold, b) by incorporating a new set of intricate driving scenarios derived from the NHTSA typology \cite{Najm2007}, and c) by increasing the frequency of these scenarios along each route. 
Additionally, this new benchmark introduces larger and more complex environments, as exemplified by Town12 and Town13. Town 12 is a $10\times10$ km$^2$ map that features a mix of urban, residential, and rural areas, offering varied types of challenges. Town13, while sharing many characteristics with Town12, distinguishes itself with different architectural styles, road and pavement textures, and vegetation types.
These enhancements aim to rigorously test the adaptability and resilience of autonomous driving systems under varied and challenging conditions.

The benchmark uses different metrics to assess different aspects of driving performance. The \acrfull{rc} indicates the percentage of the route completed by the agent. The \acrfull{ip} quantifies the severity of infractions and is calculated using the following formula:

\begin{equation}
\textbf{IP} = \prod_{i=1}^{q} p_{i}^{n_i},
\end{equation}

\noindent where $q$ denotes the total number of different infraction types, $p_i$ is the penalty associated with the infraction type $i$, and $n_i$ is the number of infractions of type $i$. The main metric of the benchmark, \acrfull{ds}, is calculated by multiplying \acrshort{rc} and \acrshort{ip}, providing a composite score that reflects both route completion success and adherence to driving regulations.

\begin{table}
\centering
\caption{Comparison of run time inference using the expert agents.}
\begin{tabular}{cc}
\toprule
          & \multicolumn{1}{c}{Run Time  $\downarrow$} \\ 

            & \si{s} \\     \midrule
Autopilot & \textbf{0.007}                            \\
PRIBOOT   & 0.011                            \\ \bottomrule
\end{tabular}
\label{tab:runtime}
\end{table}

\begin{table}
\centering
\caption{Abbrevation and the corresponding full name of the metrics used in Leaderboard 2.0.}
\begin{tabular}{cc}
\toprule
Abbreviation & Full Name \\ \midrule
\textbf{DS}           & Driving Score                 \\
\textbf{IRS}          & Infraction Rate Score         \\
\textbf{RC}           & Route Completion              \\
\textbf{IP}           & Infraction Penalty            \\
\textbf{C.P}          & Collisions Pedestrians        \\
\textbf{C.V}          & Collisions Vehicles           \\
\textbf{C.L}          & Collisions Layout             \\
\textbf{R.L}          & Red Light Infractions         \\
\textbf{Stop}         & Stop Sign Infractions         \\
\textbf{O.R}          & Off-road Infractions          \\
\textbf{R.D}          & Route Deviation               \\
\textbf{Block}        & Agent Blocked                 \\
\textbf{Y.E}          & Yield Emergency Infractions   \\
\textbf{S.T}          & Scenario Timeouts             \\
\textbf{M.S}          & Min Speed Infractions        \\ \bottomrule
\end{tabular}
\label{tab:metrics}
\end{table}

\subsubsection{\acrlong{irs}}

While \acrshort{ds} provides valuable insights into agent performance, it inherently biases against longer routes due to its cumulative penalty for infractions, which are statistically more likely to occur over extended distances. To address this discrepancy and promote fairness, we introduce the \acrfull{irs}. This metric accounts for the infraction rate per kilometer, adjusting for route length and providing a balanced evaluation across varying driving conditions. The \acrshort{irs} is defined as:

\begin{equation}
    \textbf{IRS} =  \textbf{RC} \cdot  \prod_{i=1}^{q} e^{-\lambda \cdot \frac{n_i}{L} \cdot (1 - p_i)} ,
\end{equation}

\noindent where $L$ represents the length of the route in kilometers, and $\lambda$ is a tunable exponent set to $4$ based on empirical testing to optimize the metric's sensitivity to infractions per distance traveled.

\subsubsection{Training Details}

We utilized the human driving logs provided by CARLA to train PRIBOOT. These logs correspond to 10 routes in Town12 and 10 routes in Town13 and amount to approximately 700,000 samples collected at a frequency of 20Hz. Each sample contains all the information required at each training step, including the \acrshort{bev} image, the vector of vehicle measurements, and the global location of the agent on the map. For our experiments, we used CARLA version 0.9.15. PRIBOOT was trained on a single NVIDIA A40 GPU.
During the training phase, we used a batch size of 256 and the Adam optimizer \cite{diederik2014adam} with a learning rate of 0.0001.

\subsection{Comparative Analysis}

This section outlines a comparative analysis conducted on Leaderboard 2.0, focusing exclusively on two expert agents: Autopilot and PRIBOOT. An attempt was made to adapt the Roach system to this benchmark; however, it was unsuccessful. The benchmark currently cannot support running a \acrshort{rl} algorithm like Roach due to memory leaks that prevent the execution of millions of steps without causing server crashes.

Table \ref{tab:runtime} provides a comparison of the runtime between Autopilot and PRIBOOT. In this evaluation, Autopilot achieves a runtime of 0.007 seconds, while PRIBOOT records a runtime of 0.011 seconds. This difference in performance is expected, given that Autopilot operates based on a predefined set of rules, whereas PRIBOOT processes high-dimensional inputs.

For the Leaderboard 2.0 results, 15 metrics were utilized to assess the performance of the models. These metrics are detailed in Table \ref{tab:metrics}, where each abbreviation is associated with its full metric name.

\setlength{\tabcolsep}{4.4pt}
\begin{table*}[]
\centering
\caption{Driving performance and infraction analysis of expert agents on CARLA Leaderboard 2.0 in Town12 and Town13.}
\begin{tabular}{cccccccccccccccccc}
\toprule
\multicolumn{2}{l}{\multirow{2}{*}{}}                   & \textbf{DS} $\uparrow$ & \textbf{IRS} $\uparrow$ & \textbf{RC} $\uparrow$ & \textbf{IP} $\uparrow$ & \textbf{C.P} $\downarrow$ & \textbf{C.V} $\downarrow$ & \textbf{C.L} $\downarrow$ & \textbf{R.L} $\downarrow$ & \textbf{Stop} $\downarrow$ &  \textbf{O.R} $\downarrow$  & \textbf{R.D} $\downarrow$ & \textbf{Block} $\downarrow$ & \textbf{Y.E} $\downarrow$ & \textbf{S.T} $\downarrow$ & \textbf{M.S} $\downarrow$ \\    %\cline{3-17} 
\multicolumn{2}{l}{}                                    & \si{\percent}          &    \si{\percent}                   &        \si{\percent}            &          \si{\percent}           &          \#/Km          &      \#/Km           &          \#/Km             &      \#/Km                &      \#/Km                &           \#/Km             &         \#/Km         &      \#/Km                                                              &               \#/Km       &               \#/Km  &               \#/Km                                                     \\ \midrule 
\multicolumn{1}{l}{\multirow{2}{*}{Town12}} & Autopilot &       1.22       &   0.51                 &      5.97         &    0.26             &        1.26             &     4.59          &     0.58               &         0.11                 &   1.84          &  0.62       &        0.66             &     1.26          &      \textbf{0.00}     &      0.34       & \textbf{0.00}                                                                            &                                                                 \\ %\cline{2-17} 
\multicolumn{1}{l}{}                & PRIBOOT   &      \textbf{22.80}        &         \textbf{42.75}              &        \textbf{76.46}              &      \textbf{0.30}                  &        \textbf{0.00}            &      \textbf{0.31}             &           \textbf{0.06}            &    \textbf{0.01}                   &        \textbf{0.02}              &         \textbf{0.05}         &       \textbf{0.00}                       &    \textbf{0.06}                                                                   &      0.04             &            \textbf{0.03}  & 0.11                                                     \\ \midrule
\multicolumn{1}{l}{\multirow{2}{*}{Town13}} & Autopilot &      0.99         &      0.22                 &     5.55              &       0.20                 &         0.83            &        3.06           &      0.83                 &              \textbf{0.00}         &          0.02            &        0.35          &          0.69                     &     0.69                                                                  &          \textbf{0.00}         &         0.10 & \textbf{0.00}                                                        \\ %\cline{2-17} 
\multicolumn{1}{l}{}                        & PRIBOOT   &      \textbf{18.84}         &       \textbf{46.97}                &         \textbf{74.29}          &      \textbf{0.24}                  &        \textbf{0.01}             &    \textbf{0.34}               &        \textbf{0.05}               &     \textbf{0.00}                  &     \textbf{0.01}                 &       \textbf{0.05}           &     \textbf{0.00}           &        \textbf{0.04}                                     &     0.02                                &      \textbf{0.02}             &      0.06                                                           \\ \bottomrule
\end{tabular}
\label{tab:comparative}
\end{table*}

\begin{figure*}[]
    \centering
    \subfloat[Autopilot: $t=0 \si{s}$]{\includegraphics[width=0.20\textwidth]{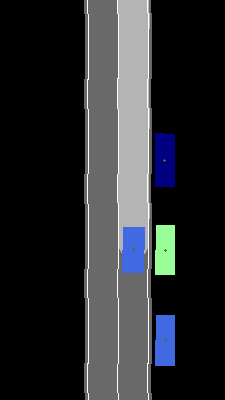}}\hspace{5pt}
    \subfloat[Autopilot: $t=2 \si{s}$]{\includegraphics[width=0.20\textwidth]{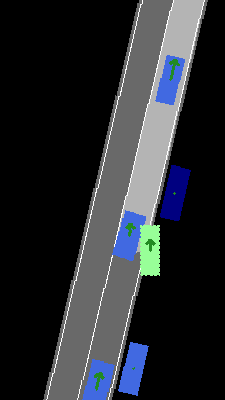}}\hspace{5pt}
    \subfloat[Autopilot: $t=3 \si{s}$]{\includegraphics[width=0.20\textwidth]{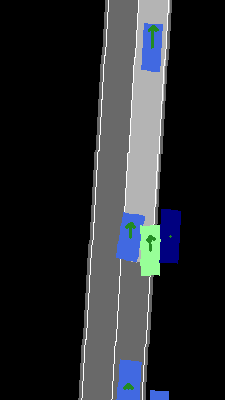}}\hspace{5pt}
    \subfloat[Autopilot: $t=4 \si{s}$]{\includegraphics[width=0.20\textwidth]{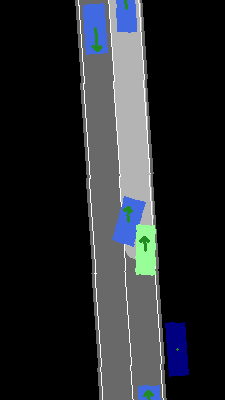}}\hspace{5pt}
    % \caption*{Sequence of frames showing AutoPilot's performance in obstacle avoidance.}
    
    \subfloat[PRIBOOT: $t=0 \si{s}$]{\includegraphics[width=0.20\textwidth]{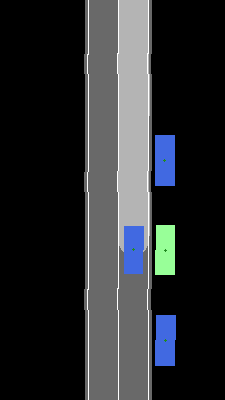}}\hspace{5pt}
    \subfloat[PRIBOOT: $t=8 \si{s}$]{\includegraphics[width=0.20\textwidth]{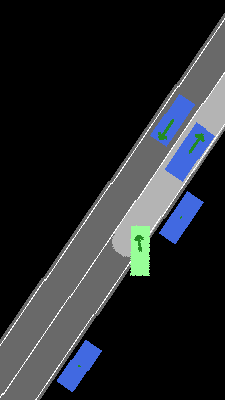}}\hspace{5pt}
    \subfloat[PRIBOOT: $t=9 \si{s}$]{\includegraphics[width=0.20\textwidth]{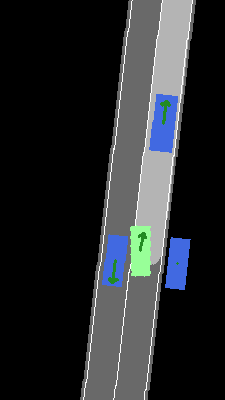}}\hspace{5pt}
    \subfloat[PRIBOOT: $t=10 \si{s}$]{\includegraphics[width=0.20\textwidth]{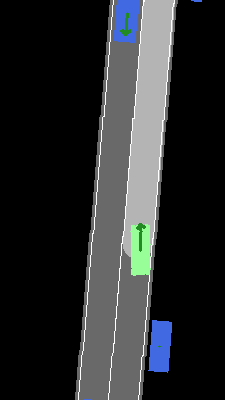}}\hspace{5pt}
    % \caption*{Sequence of frames showing AutoPilot's performance in obstacle avoidance.}
    
    \caption{Qualitative comparison in a parking exit scenario between Autopilot and PRIBOOT. The first row depicts a sequence of keyframes from Autopilot, while the second row shows the keyframes from PRIBOOT.}

    \label{fig:exiting_park}
\end{figure*}

\begin{figure*}[h!]
    \centering
    \subfloat[Autopilot: $t=0 \si{s}$]{\includegraphics[width=0.20\textwidth]{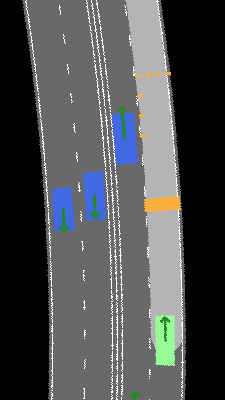}}\hspace{5pt}
    \subfloat[Autopilot: $t=2 \si{s}$]{\includegraphics[width=0.20\textwidth]{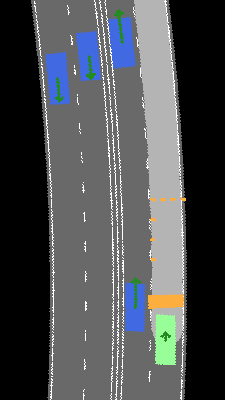}}\hspace{5pt}
    \subfloat[Autopilot: $t=4 \si{s}$]{\includegraphics[width=0.20\textwidth]{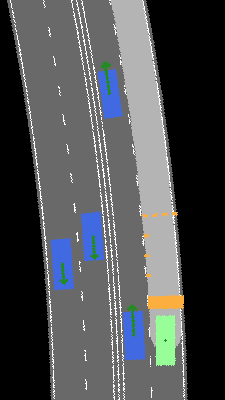}}\hspace{5pt}
    \subfloat[Autopilot: $t=6 \si{s}$]{\includegraphics[width=0.20\textwidth]{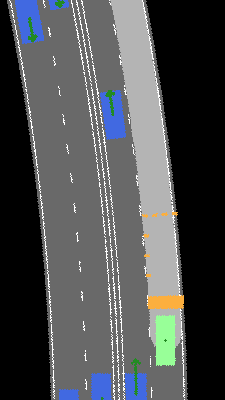}}\hspace{5pt}
    % \caption*{Sequence of frames showing AutoPilot's performance in obstacle avoidance.}
    
    \subfloat[PRIBOOT: $t=0 \si{s}$]{\includegraphics[width=0.20\textwidth]{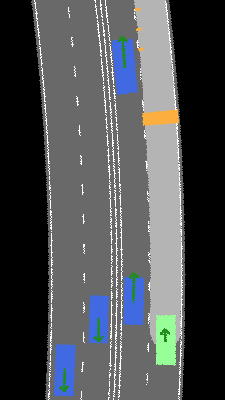}}\hspace{5pt}
    \subfloat[PRIBOOT: $t=2 \si{s}$]{\includegraphics[width=0.20\textwidth]{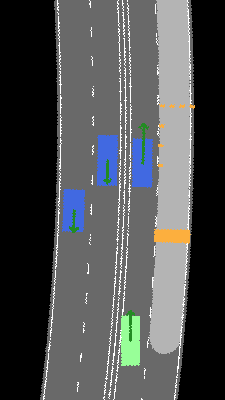}}\hspace{5pt}
    \subfloat[PRIBOOT: $t=4 \si{s}$]{\includegraphics[width=0.20\textwidth]{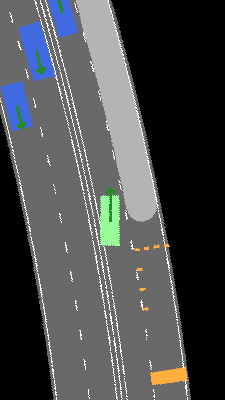}}\hspace{5pt}
    \subfloat[PRIBOOT: $t=6 \si{s}$]{\includegraphics[width=0.20\textwidth]{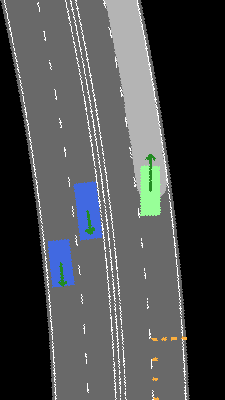}}\hspace{5pt}
    % \caption*{Sequence of frames showing AutoPilot's performance in obstacle avoidance.}
    
    \caption{Qualitative comparison in a lane obstacle scenario between Autopilot and PRIBOOT. The first row depicts a sequence of keyframes from Autopilot, while the second row shows the keyframes from PRIBOOT.}
    \label{fig:obstacle_avoidance}
\end{figure*}

As demonstrated in Table \ref{tab:comparative}, we conducted performance comparisons of the agents in two distinct Towns: Town12 and Town13. The evaluations are based on averages derived from 90 routes in Town12 and 20 routes in Town13, as stipulated in Leaderboard 2.0. PRIBOOT consistently outperformed Autopilot across nearly all metrics in both towns, often by substantial margins. In Town12, for instance, PRIBOOT's \acrshort{ds} was approximately 19 times higher than that of Autopilot, and its \acrshort{irs} was 84 times better. Similar trends were observed in Town13, with PRIBOOT achieving 19 times higher \acrshort{ds} and 214 times higher \acrshort{irs} than Autopilot. Notably, PRIBOOT recorded zero collisions with pedestrians per kilometer in Town12 and only 0.01 collisions per kilometer in Town13, underscoring its effectiveness in minimizing accidents involving pedestrians.

In contrast, Autopilot demonstrated superior performance in two specific metrics: yielding to emergency vehicles and maintaining minimum speed. The former was due to its lower \acrshort{rc} score, which resulted in zero scenarios requiring yielding to an emergency vehicle. The latter stems from Autopilot operating at a fixed target speed consistently above the minimum speed requirement for the roads where the agent drove.

PRIBOOT stands out as the first agent to achieve a \acrshort{rc} of approximately 75\% in both towns, coupled with a satisfactory \acrshort{ds} and \acrshort{irs}. This marks a significant milestone, positioning PRIBOOT as a pioneering agent capable of navigating the complexities of the benchmark, which can be used for data collection or online demonstrations.

% Qualitative comparisons

To enhance the quantitative comparison presented earlier, we also include a qualitative evaluation. Our analysis of all routes in the benchmark reveals that Autopilot struggles with the novel scenarios introduced by Leaderboard 2.0, particularly those requiring slight deviations from the global planner's trajectory. These scenarios include instances like parking exits and lane obstacles.
Figure \ref{fig:exiting_park} illustrates a sequence of keyframes in a parking exit scenario, first showing Autopilot's performance and then PRIBOOT's. As shown, Autopilot immediately exits the park without considering the vehicles in the lane, leading to a collision. Conversely, PRIBOOT waits for a moment when the lane is clear before exiting, as expected.
Figure \ref{fig:obstacle_avoidance} also shows a sequence of keyframes, this time involving an obstacle in the lane. Here, Autopilot approaches the obstacle and then stops, whereas PRIBOOT slightly deviates from the trajectory to avoid the obstacle and returns to the original path once it is clear.
These qualitative comparisons demonstrate that PRIBOOT is better equipped to handle the challenging new driving scenarios introduced by Leaderboard 2.0.

Additionally, we provide access to a series of demonstration videos that illustrate the performance of PRIBOOT on Leaderboard 2.0. These can be accessed \href{https://drive.google.com/drive/folders/1NJa4bSQ-pptq1OFHyDRWweQVSnbfxPc3?usp=sharing}{here}.

\subsection{Ablation Study}

To explore the individual contributions of key components within PRIBOOT, particularly under conditions of limited data, we performed an ablation study focusing on two crucial elements: data augmentation and the utilization of RGB \acrshort{bev} in conjunction with transfer learning. This study involved training two variants of PRIBOOT: the first variant (referred to as "w/o aug") was developed without the data augmentations depicted in Figure \ref{fig:aug}, and the second variant (referred to as "w/ masks") employed the \acrshort{bev} as a set of masks and training a \acrshort{cnn} from scratch, consistent with methodologies reported in the literature \cite{zhang2021end}.

The comparative analysis of driving performance and infractions for these variants is presented in Table \ref{tab:ablation}. The w/o aug variant exhibited a significant decline in performance, as evidenced by a \textbf{RC} of approximately 16\%, which adversely affected all other performance metrics. The reason for this is that due to planning or control inaccuracies, the agent encounters situations where it deviates from the center of the lane and lacks the capability to effectively recover. On the other hand, the w/ masks variant demonstrated improved results compared to the w/o aug variant. However, it still fell short of the full PRIBOOT system's capabilities. Specifically, the w/ masks variant scored three times lower in terms of \textbf{DS} and two times lower in terms of \textbf{IRS}.

\setlength{\tabcolsep}{4.4pt}
\begin{table*}
\centering
\caption{Ablation Study: Driving performance and infraction analysis of PRIBOOT variants on CARLA Leaderboard 2.0 in Town13.}
\begin{tabular}{cccccccccccccccccc}
\toprule
\multicolumn{2}{l}{\multirow{2}{*}{}}                   & \textbf{DS} $\uparrow$ & \textbf{IRS} $\uparrow$ & \textbf{RC} $\uparrow$ & \textbf{IP} $\uparrow$ & \textbf{C.P} $\downarrow$ & \textbf{C.V} $\downarrow$ & \textbf{C.L} $\downarrow$ & \textbf{R.L} $\downarrow$ & \textbf{Stop} $\downarrow$ &  \textbf{O.R} $\downarrow$  & \textbf{R.D} $\downarrow$ & \textbf{Block} $\downarrow$ & \textbf{Y.E} $\downarrow$ & \textbf{S.T} $\downarrow$ & \textbf{M.S} $\downarrow$ \\    %\cline{3-17} 
\multicolumn{2}{l}{}                                    & \si{\percent}          &    \si{\percent}                   &        \si{\percent}            &          \si{\percent}           &          \#/Km          &      \#/Km           &          \#/Km             &      \#/Km                &      \#/Km                &           \#/Km             &         \#/Km         &      \#/Km                                                              &               \#/Km       &               \#/Km  &               \#/Km                                                     \\ \midrule 

\multicolumn{1}{l}{\multirow{3}{*}{Town13}} & w/o aug &     2.66          &        5.22             &     15.92          &         \textbf{0.29}               &      \textbf{0.01}             &      1.22            &     0.35                &    \textbf{0.00}                  &       0.02               &          0.45        &       0.02                  &     0.40                                                                &        \textbf{0.00}       &    0.17      &     \textbf{0.02}                                             \\ %\cline{2-17} 

\multicolumn{1}{l}{}                        & w/ masks   &    5.55         &     21.08                  &       54.02          &     0.18            &    0.05       &    0.52      &      0.23               &       0.01             &    \textbf{0.00}             &      0.08     &     0.03       &    0.06                                    &       0.05                      &    0.11               &   0.14    \\

\multicolumn{1}{l}{}                        & PRIBOOT   &      \textbf{18.84}         &       \textbf{46.97}                &         \textbf{74.29}          &      0.24                 &        \textbf{0.01}             &    \textbf{0.34}               &        \textbf{0.05}               &     \textbf{0.00}                  &     0.01                 &       \textbf{0.05}           &     \textbf{0.00}           &        \textbf{0.04}                                     &     0.02                               &      \textbf{0.02}             &      0.06                                                          \\ \bottomrule
\end{tabular}
\label{tab:ablation}
\end{table*}

\begin{figure}[]
% \includesvg[width=2\columnwidth]{imgs/iccv_2023_network_v3}
\includegraphics[width=1.0\columnwidth]{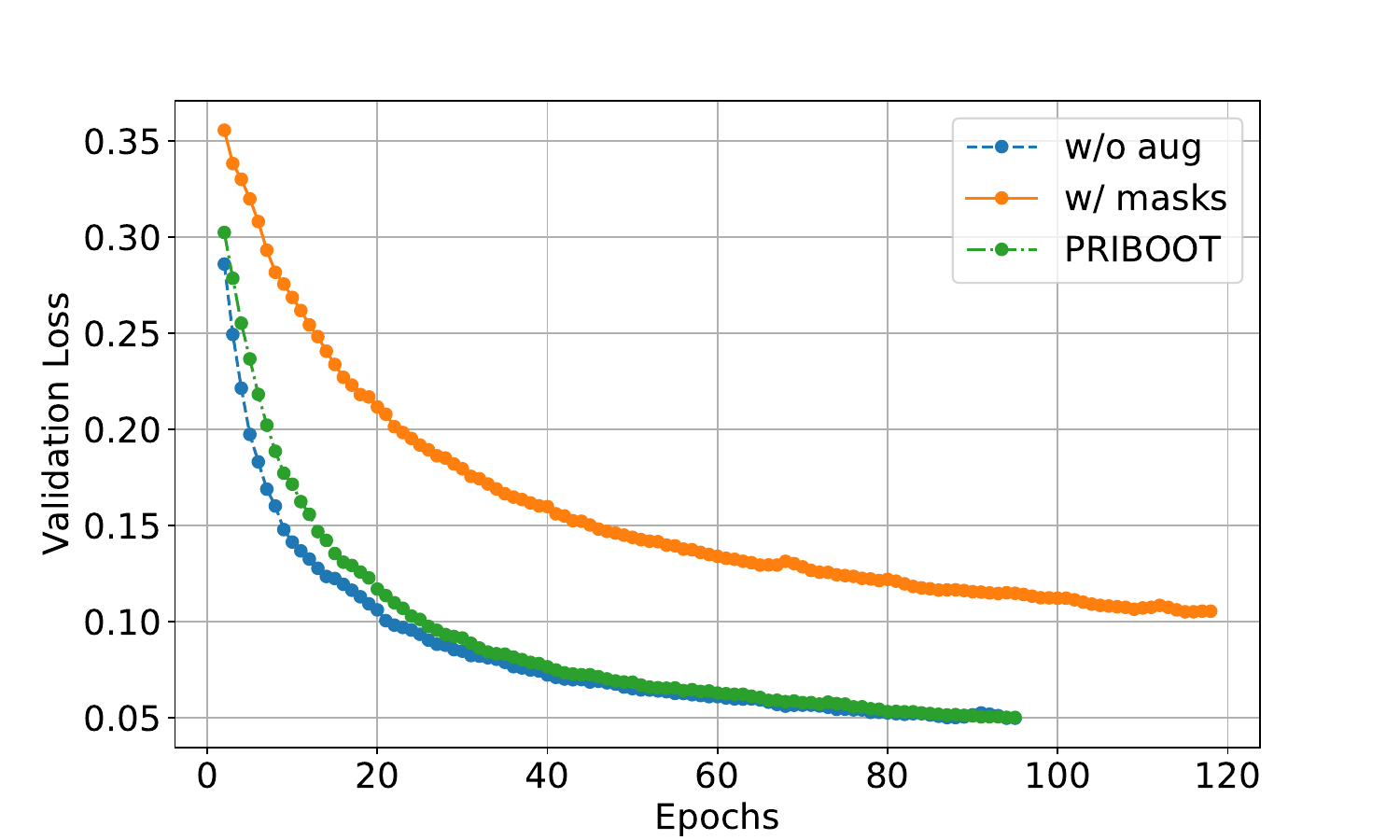}
\caption{Validation loss across epochs of PRIBOOT variants.}
\label{fig:val_loss}
\end{figure}

Figure \ref{fig:val_loss} illustrates the validation loss across epochs for the ablations considered. While the w/o aug variant achieves results similar to PRIBOOT during training, it recurrently faces out-of-distribution events, as detailed in Table \ref{tab:ablation}. In contrast, the w/ masks variant displays a distinct pattern in validation loss: it requires 20\% more epochs to converge and converges at a loss value that is twice that of PRIBOOT. This performance deficit underscores the critical role of utilizing RGB \acrshort{bev} and transfer learning techniques in cases where data availability is limited.

%--------------------------------------------------------------------------------------
%--------------------------------------------------------------------------------------
\section{Conclusion}
%--------------------------------------------------------------------------------------
%--------------------------------------------------------------------------------------

In this paper, we introduced PRIBOOT, a system that utilizes privileged information alongside limited human driving logs to establish the first expert with satisfactory driving performance on the CARLA Leaderboard 2.0. Our results demonstrate that PRIBOOT significantly outperforms Autopilot across nearly all benchmark metrics, highlighting its superior capability in complex and challenging autonomous driving scenarios.
Additionally, we presented an ablation study that evaluates the impact of using augmentations to aid recovery processes. Furthermore, we demonstrated the benefits of employing RGB \acrshort{bev} images with transfer learning, which proved more efficient in terms of training speed and performance than using masks and training a CNN from scratch. While our work has focused on the CARLA simulator, it is important to note that the idea behind PRIBOOT can eventually be applied in other simulators.
In the future, we plan to employ PRIBOOT to generate large datasets that can be instrumental in training student models that receive sensor information as input.

\section*{Acknowledgements}
This work has been supported by FCT - Foundation for Science and
Technology, in the context of Ph.D. scholarship 2022.10977.BD and by National Funds through the FCT - Foundation for Science and Technology, in the context of the project UIDB/00127/2020.

{\small
\bibliographystyle{IEEEtran}
% \bibliography{egbib}

}

\vfill

\end{document}